\documentclass[letterpaper, 10 pt, conference]{ieeeconf}  

\IEEEoverridecommandlockouts                              

\usepackage{times}
\usepackage{epsfig}
\usepackage{graphicx}
\usepackage{amsmath}
\usepackage{amssymb}
\usepackage[font=footnotesize]{caption}
\usepackage{subcaption}
\usepackage{siunitx}
\usepackage{hyperref}
\usepackage{xcolor}
\newcommand{\modelname}{Residual-NeRF}

\title{\LARGE \bf
\modelname{}: Learning Residual NeRFs \\ for Transparent Object Manipulation
}

\author{Bardienus P. Duisterhof$^{1}$, Yuemin Mao$^{1}$, Si Heng Teng$^{1}$,
 Jeffrey Ichnowski$^{1}$
\thanks{$^{1}$Carnegie Mellon University, The Robotics Institute, {\tt\small \{bduister, jeffi\}@cmu.edu}
        }%
}

\begin{document}

\maketitle
\thispagestyle{empty}
\pagestyle{empty}

\begin{abstract}
Transparent objects are ubiquitous in industry, pharmaceuticals, and households. Grasping and manipulating these objects is a significant challenge for robots. Existing methods have difficulty reconstructing complete depth maps for challenging transparent objects, leaving holes in the depth reconstruction. Recent work has shown neural radiance fields (NeRFs) work well for depth perception in scenes with transparent objects, and these depth maps can be used to grasp transparent objects with high accuracy. NeRF-based depth reconstruction can still struggle with especially challenging transparent objects and lighting conditions.
In this work, we propose \modelname{}, a method to improve depth perception and training speed for transparent objects.
Robots often operate in the same area, such as a kitchen. By first learning a \emph{background NeRF} of the scene without transparent objects to be manipulated, we reduce the ambiguity faced by learning the changes with the new object. We propose training two additional networks: a \emph{residual NeRF} learns to infer residual RGB values and densities, and a \emph{Mixnet} learns how to combine background and residual NeRFs. We contribute synthetic and real experiments that suggest \modelname{} improves depth perception of transparent objects. The results on synthetic data suggest \modelname{} outperforms the baselines with a 46.1 \% lower RMSE and a 29.5 \% lower MAE. Real-world qualitative experiments suggest \modelname{} leads to more robust depth maps with less noise and fewer holes.
Website: \url{https://residual-nerf.github.io}

\end{abstract}

\section{Introduction}
Dexterous manipulation of transparent objects can enable robots to perform new and impactful tasks in industry, pharmaceuticals, and households. Robots often use depth images of objects to decide what action (e.g., pull, lift, or drop) to perform. However, common depth sensors struggle to capture depth images for arbitrary transparent objects~\cite{IchnowskiAvigal2021DexNeRF,chen2022clearpose,glassware_learning,s20236790}. Learning-based approaches for transparent object depth estimation work well in-distribution, but can struggle to generalize outside their training data~\cite{IchnowskiAvigal2021DexNeRF}. The lack of surface features on transparent objects also makes it challenging to retrieve depth maps using approaches such as COLMAP~\cite{schoenberger2016sfm}.

Neural radiance fields (NeRFs)~\cite{mildenhall2020nerf} are implicit neural network scene representations trained on multiple views of the same scene and capable of state-of-the-art novel view synthesis. NeRF combines a multi-layer perception (MLP) with classic volumetric rendering techniques to achieve photorealistic novel view synthesis results. The only assumptions in NeRF are those imposed by traditional volumetric rendering techniques~\cite{mildenhall2020nerf}. 

Recently, Dex-NeRF~\cite{IchnowskiAvigal2021DexNeRF} and Evo-NeRF~\cite{pmlr-v205-kerr23a} showed that NeRFs are effective at depth perception of transparent objects. However, these methods also showed that NeRFs tend to struggle with particularly challenging transparent objects, such as wine glasses or kitchen foil with challenging light conditions. The challenge with transparent objects originates from a lack of features, and large view-dependent changes in appearance.

Dex-NeRF, while achieving high grasp success rates, was slow to compute.  Evo-NeRF~\cite{kerr2022evo} significantly sped up NeRF-based grasping by \emph{evolving} NeRFs. In this work, we extend on the Evo-NeRF motivations: (1) speed up training to a successful grasp, and (2) learning NeRFs for changing scenes, where small gradients in empty regions can be challenging~\cite{kerr2022evo}.


\begin{figure}
    \centering
    \includegraphics[width = \linewidth]{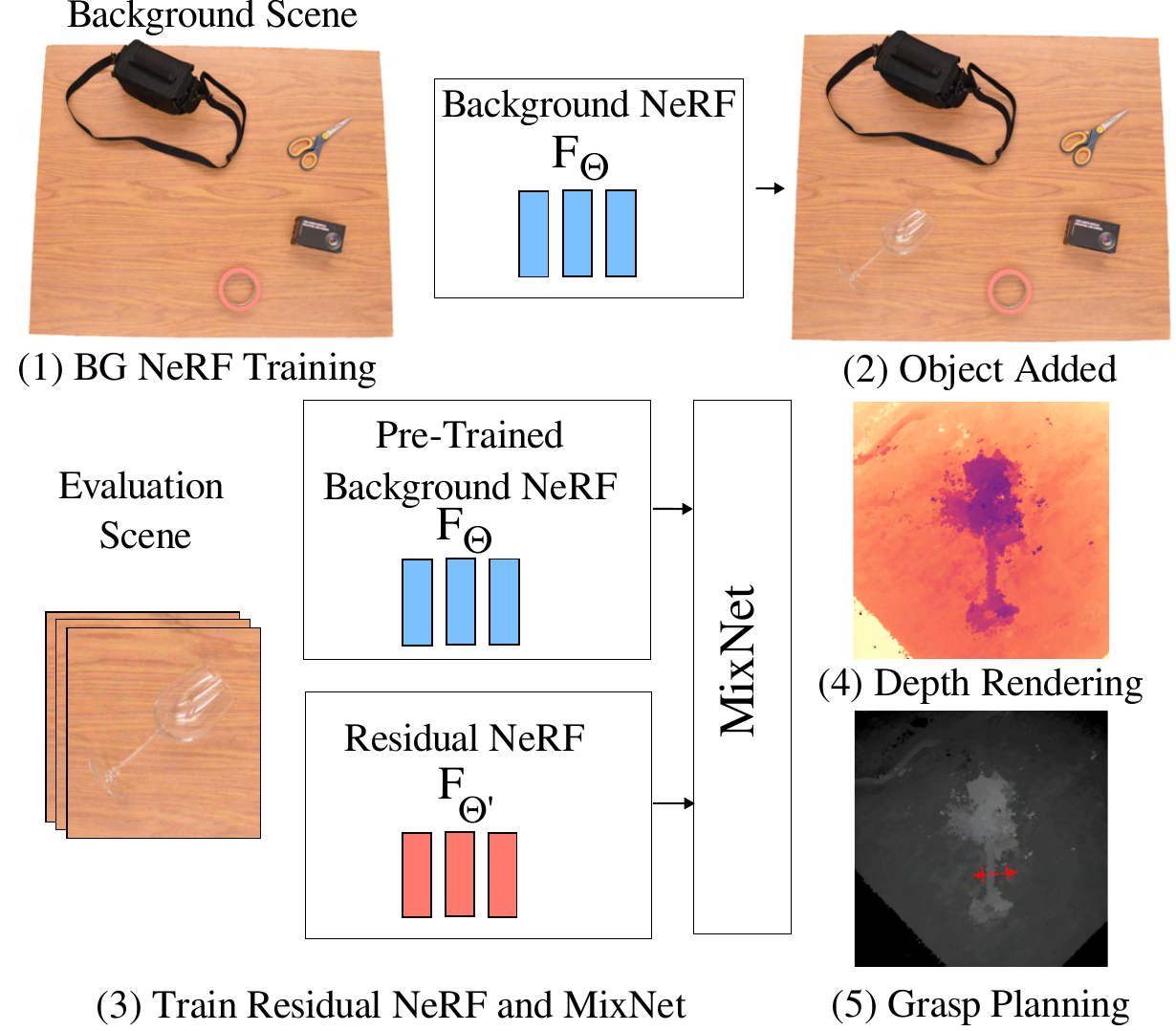}
    \caption{\modelname{}, a method that leverages mostly static scenes to improve depth perception and speed up training. \modelname{} begins by learning a \emph{background NeRF} of the entire scene without transparent objects. Following this, we learn a \emph{residual NeRF} and a \emph{Mixnet} to complement the \emph{background NeRF}. }
    \label{fig:fig_1}
\end{figure}


We propose \modelname{} (Figure~\ref{fig:fig_1}), a method to leverage a strong scene prior to improving depth perception of transparent objects. In many scenarios, the geometry of the robot's work area is mostly static and opaque, e.g., shelves, desks, and tables. \modelname{} leverages the static and opaque parts of the scene as a prior, to reduce ambiguity and improve depth perception. \modelname{} first learns a \emph{background NeRF} of the entire scene by training on images without the transparent objects present. \modelname{} then uses images of the full scene with the transparent objects to learn a \emph{residual NeRF} and a \emph{Mixnet}. The algorithm adaptively mixes both NeRFs for each point along the ray, using a novel \emph{Mixnet} multi-layer perceptron (MLP). The Mixnet infers a mixing weight between the background and residual NeRF using a location-dependent multi-resolution hash encoding. 

We evaluate \modelname{} on nine photo-realistic synthetic and three real scenes and compare its performance to other relevant NeRF algorithms. We compare 1) learning speed, 2) depth reconstruction, and 3) robot grasp quality. The results suggest that \modelname{} improves on the state-of-the-art in depth mapping accuracy with a 46.1 \% lower RMSE and a 29.5 \% lower MAE. The results also suggest \modelname{} speeds up training and leads to more robust grasp planning.

\begin{figure*}[t]
     \centering
     \includegraphics[width=\textwidth]{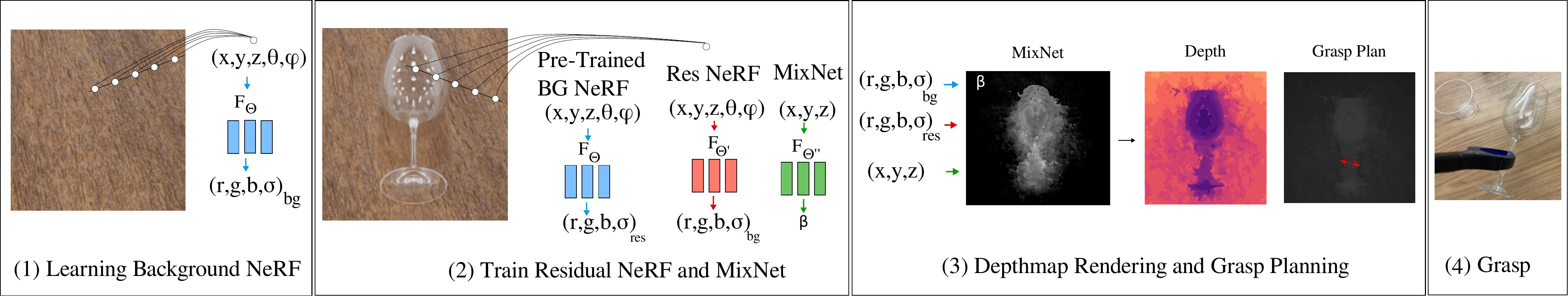}
     \caption{Residual-NeRF, a method that leverages mostly static scenes to
improve depth perception and speed up training. We first learn a \emph{background NeRF} of the scene without transparent objects and leverage it as a scene prior. Following this, we learn a \emph{residual NeRF} and \emph{Mixnet}. The \emph{Mixnet} is an MLP that learns to combine the \emph{background NeRF} and the \emph{residual NeRF}. Equation~\ref{eq:residual_nerf} describes how the output of the \emph{Mixnet} MLP is used to combine the two NeRFs.}
\vspace{-5mm}
\label{fig:method}

\end{figure*}

In summary, we contribute:
\begin{itemize}
    \item \modelname{}, a novel algorithm that learns a \emph{residual NeRF} and \emph{background NeRF}, mixed in 3D space by a \emph{Mixnet} MLP.
    \item Experiments that suggest \modelname{} infers improved depth maps of transparent objects, speeds up training, and yields more robust grasp planning. 
    \item Synthetic and real datasets.
\end{itemize}


\section{Related Work}
\label{sec:related_work}


\subsubsection*{\bf Neural Radiance Fields (NeRF)} NeRF~\cite{mildenhall2020nerf} is a scene representation capable of photo-realistic novel view reconstruction. It is inspired by classic volumetric rendering techniques, tracing rays through the scene to infer radiance and density using a neural network. For every point in 3D space, the NeRF MLP infers RGB radiance and density to render an RGB image. NeRF forms the basis for sensing in this paper.

After the introduction of NeRF, several works proposed improvements. A key challenge for the original NeRF algorithm~\cite{mildenhall2020nerf} was the long training and inference time---often requiring several hours to train for a single scene. Recent progress has shown that novel representations and system optimizations~\cite{mueller2022instant,DBLP:journals/corr/abs-2103-13744,liu2020neural,yu2021plenoctrees,SunSC22,9710021,10.1145/3450626.3459863,mueller2022instant,mubarik2023hardware}, and depth supervision~\cite{9880067,attal2021torf,wei2021nerfingmvs,neff2021donerf,Sucar:etal:ICCV2021} can significantly speed up training. We leverage existing optimized code to speed up training~\cite{torch-ngp}. 

Other works have focused on improving novel view synthesis performance in challenging conditions and with sparse camera views without extrinsic calibration. It is now possible to capture a NeRF of transparent objects~\cite{pmlr-v205-kerr23a,IchnowskiAvigal2021DexNeRF}, reflective surfaces~\cite{verbin2022refnerf} and low-light environments~\cite{mildenhall2022rawnerf}. Many NeRF and other works have aimed to reduce the number of views~\cite{zhang2021ners,SRF,sparf2023,Niemeyer2021Regnerf,zhang2024raydiffusion} and extrinsic camera calibration~\cite{lin2021barf,yen2020inerf,chen2023local,Jeong_2021_ICCV,zhang2024raydiffusion} required. Another line of work extends novel view synthesis to dynamic scenes~\cite{TiNeuVox,duisterhof2023mdsplatting,wu20234dgaussians,luiten2023dynamic}.

\subsubsection*{\bf Merging NeRFs} 
Previous work has introduced methods to enable evaluating and training multiple NeRFs in a single scene. Reiser et al.~\cite{reiser2021kilonerf} introduced KiloNeRF, with the contribution of using thousands of smaller MLPs ordered in a 3D grid instead of using only a single MLP. KiloNeRF maps a position $\mathbf{x}$ to a tiny NeRF using spatial binning to its 3D grid, and achieves faster inference.

Tancik et al.~\cite{tancik2022blocknerf} introduced Block-NeRF, a variant of NeRF that can represent large-scale environments. Block-NeRF also contains an appearance embedding to accommodate changes in, e.g., lighting. Zhang et al.~\cite{zhang2022nerfusion} introduced NeRFusion, a method that leverages TSDF-based fusion techniques for efficient large-scale reconstruction. The authors propose to infer a per-frame local radiance field via direct network inference. They are then fused using a recurrent neural network to reconstruct a global scene representation.

In summary, related work has demonstrated novel approaches to improve view reconstruction and inference speed in large scenes. \modelname{} aims to improve depth mapping of transparent objects for manipulation, by learning a Residual NeRF and Mixnet on top of a pretrained background NeRF. Mixnet does not require a pre-defined area for each NeRF, but learns to mix both NeRFs at every point in space.

\subsubsection*{\bf Depth Perception of Transparent Objects}
Several works have proposed methods for accurate depth perception, shape estimation, and/or pose estimation. Xie et al.~\cite{transformer_transparent} developed a pipeline based on transformer neural networks capable of transparent object segmentation. Phillips et al.~\cite{glassware_learning} leveraged a random forest algorithm to extract the pose and shape of transparent objects. Xu et al.~\cite{s20236790} contributed an algorithm for estimating 6-degrees-of-freedom (DOF) pose of a transparent object using only a single RGBD image. Wang et al.~\cite{wang2023mvtrans} contributed MVTrans for depth mapping, segmentation, and pose estimation of transparent objects. Chen et al.~\cite{chen2022clearpose} contributed a benchmark dataset for segmentation, object-centric pose estimation, and depth completion. 

Ichnowski et al.~\cite{IchnowskiAvigal2021DexNeRF} showed how NeRFs can be leveraged to infer state-of-the-art depth perception of transparent objects, and unlike data-centric approaches, did not require a prior training on a set of objects.

\section{Problem Statement}
Given is a sequence of images of a scene without the transparent objects present. $I_{\mathrm{bg}_n}$ with $n \in [1,\dots,N_\mathrm{bg}]$ each with camera matrix $P_n$ (i.e., the intrinsics and extrinsics). In addition, we also have access to images of the same scene with one or more transparent objects present, $I_{\mathrm{res}_n}$, with $n \in [1,\dots,N_\mathrm{res}]$, $P'_n$. $P'_n$ and $P_n$ are not necessarily the same, $N_\mathrm{bg}$ and $N_\mathrm{res}$ are also not necessarily equal. 

The objective is to recover novel view depth maps from any given (depth) camera pose $P$ in the scene, and then use the novel views for grasp planning. As with Dex-NeRF and Evo-NeRF, the goal is to facilitate using a grasp planner such as Dex-Net~\cite{DBLP:journals/corr/MahlerLNLDLOG17}. A high-quality depth map of both opaque and transparent objects in the scene will aid motion planning and the selection of grasp location. Perceived holes in objects can lead to collisions, whereas hallucinating non-existant surfaces may lead to no available grasp location.

\section{Method}
Residual-NeRF, shown in Figure~\ref{fig:method}, recovers depth using multiple camera views by first learning a background NeRF, then training a Mixnet and a residual NeRF (Section~\ref{sec:residual NeRF}). As it builds on NeRF, we first review prelimaries of novel view synthesis with NeRF (Section~\ref{sec:background}). Once trained, we then use the NeRFs to render depth maps (Section~\ref{sec:nerf_depth}) for the grasp planner.


\subsection{Preliminary: Training NeRF}
\label{sec:background}
NeRFs~\cite{mildenhall2020nerf} trace rays through images of the scene to train an MLP, $\Phi_r$, to infer radiance and density $(RGB \sigma)$ for an input world coordinate and viewing direction $(x,y,z,\theta,\phi).$  To accommodate high-frequency changes in 3D space, the input to NeRF is concatenated with a positional encoding in the form of sinusoids with varying frequencies. Some recent NeRF implementations use alternative representations such as a multi-resolution hash grid~\cite{mueller2022instant} encoding to speed up training. Thus, $\Phi_r ( h(x,y,z,\theta,\phi) ) = (RGB \sigma)$, where $h(\cdot)$ is the encoding function.

The expected color $C(\mathbf{r})$ by volumetric rendering is in Equation~\ref{eq:volumetric}. 

\begin{equation}
\label{eq:volumetric}
    C(\mathbf{r}) = \int_{t_n}^{t_f} T(t) \sigma(\mathbf{r}(t))\mathbf{c}(\mathbf{r}(t),\mathbf{d})dt,
\end{equation}
with 
\begin{equation*}
    T(t) = \exp\left(-\int_{t_n}^{t}\sigma (\mathbf{r}(s))ds\right),
\end{equation*}
$T(t)$ is the accumulated transmittance along the ray from $t_n$ to $t$. NeRFs approximate the function $C(\mathbf{r})$ by a learned function $\hat{C}(\mathbf{r})$, computed using Equation~\ref{eq:volumetric_nerf}.

\begin{equation}
\hat{C}(\mathbf{r}) = \sum_{i=1}^{N} T_i (1-\exp(-\sigma_i \delta_i))\mathbf{c}_i,
\label{eq:volumetric_nerf}
\end{equation}
with
\begin{equation*}
    T_i = \exp\left(-\sum_{j=1}^{i-1}\sigma_j \delta_j\right),
\end{equation*}

Here $\delta_i = t_{i+1} - t_i$, i.e., the distance between samples along the ray. The MLP learns $\mathbf{c}_i$, the color emitted at each point on the ray, and $\sigma_i$, the density at each point on the ray.

A NeRF can be trained by setting the loss to the RMSE between rendered and GT images. Equation~\ref{eq:photometric_loss} defines this photometric loss.
\begin{equation}
\label{eq:photometric_loss}
    \mathcal{L}_{\mathrm{pho}} = \sum_{(i,\mathbf{r})\in \Omega_r}|| \hat{C}_i(\mathbf{{r}}) - C_i(\mathbf{r}) ||_2^2,
\end{equation}
Here $i \in [1,...,N] $ is the frame number, $\mathbf{r}$ is the pixel location, and $\Omega_{r}$ is the set of all pixel locations across all frames.

\subsection{Depth from NeRF}
\label{sec:nerf_depth}
We adopt the approach for depth perception of transparent objects proposed by Ichnowski et al.~\cite{IchnowskiAvigal2021DexNeRF}. Prior works propose computing depth using Equation~\ref{eq:bad_depth},
\begin{equation}
\label{eq:bad_depth}
\hat{D}(\mathbf{r}) = \sum_{i}^{N} T_i(1-\exp(-\sigma_i \delta_i)) t_i,
\end{equation}

When propagating rays through transparent objects, this equation proves to be inaccurate as a result of multiple surfaces radiating light along the ray. Instead, as proposed by Ichnowski et al.~\cite{IchnowskiAvigal2021DexNeRF}, we find the point closest to the camera for which $\sigma_i \geq m$, where $m$ is a tunable parameter. Prior work suggested that $m$ was not overly sensitive and that a single tuned value could be reused across multiple scenes and lighting conditions.

\subsection{Learning a Residual NeRF}
\label{sec:residual NeRF}
In this work, we propose a method to leverage a NeRF of the background scene (i.e., the scene without the transparent object) to improve 3D reconstruction. Equation~\ref{eq:residual_nerf} shows the modification.

\begin{equation}
\label{eq:residual_nerf}
\resizebox{0.88\linewidth}{!}{$
\hat{C}(\mathbf{r}) = \sum_{i=1}^{N} T_i (1-\text{exp}(-((1-\beta)\sigma_\mathrm{bg}+\beta \sigma_\mathrm{res})_i \delta_i))(\mathbf{c}_{\mathrm{bg}+\mathrm{res}})_i,
$}
\end{equation}
with
\begin{align*} 
    \mathbf{c}_{\mathrm{bg}+\mathrm{res}} & = S((1.0-\beta) \mathbf{c'}_\mathrm{bg}+\beta \mathbf{c'}_\mathrm{res}), \\
    (\mathbf{c'}_\mathrm{res}, \sigma_{res}) & = \Phi_\mathrm{res} ( h(x,y,z,\theta,\phi)),  \\
    (\mathbf{c'}_\mathrm{bg}, \sigma_{bg}) & = \Phi_\mathrm{bg} ( h(x,y,z,\theta,\phi)),  \\
    \beta &= S(\Phi_\mathrm{mix} ( h(x,y,z))),  \\
    S(x) & = \frac{1}{1+e^{-x}}, 
\end{align*}

For every point along the ray, MLP $\Phi_\mathrm{mix}$ takes in the encoded position and infers $\beta \in [0,1]$. The background NeRF infers $\sigma_\mathrm{bg}$ and $\mathbf{c'}_\mathrm{bg}$ by training on a set of images of the background without the transparent object, while the residual NeRF infers $\sigma_\mathrm{res}$ and $\mathbf{c'}_\mathrm{res}$ by training with the transparent object in the scene. The vector $\mathbf{c'}$ is the color $\mathbf{c}$ before it is passed through its sigmoid activation. We add the background and residual NeRF values before going through their final activation layers, such as the sigmoid. The background NeRF is not updated while training the residual NeRF.

\begin{figure}[t]
    \includegraphics[width=\linewidth]{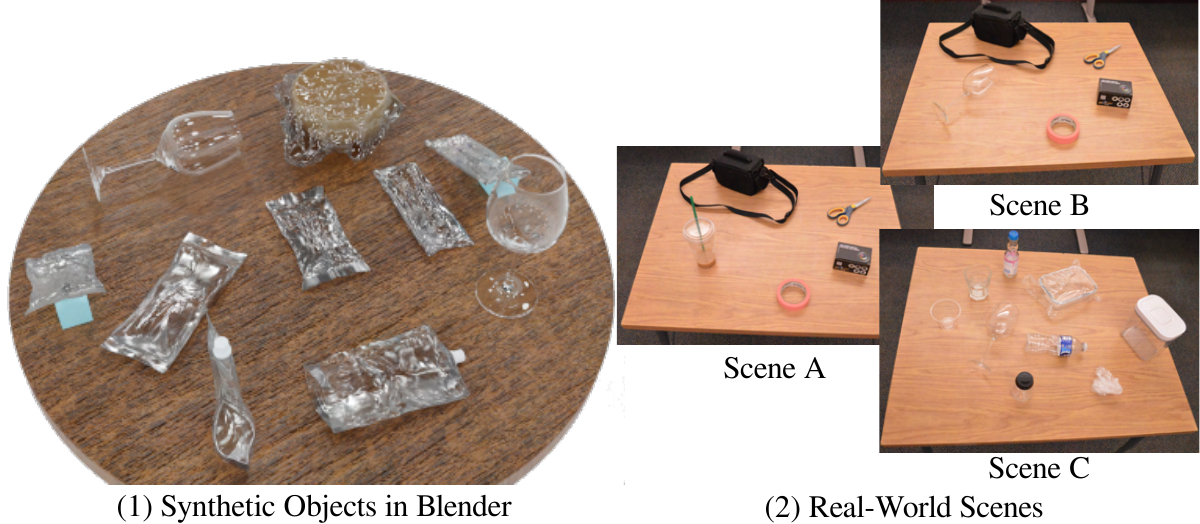}
    \caption{The scenes used for evaluation. We create nine synthetic Blender~\cite{blender} scenes with transparent objects and three real-world scenes, divided into scenes A-C with increasing difficulty. }
    \label{fig:scenes}
    \vspace{-5mm    }
\end{figure}

\begin{figure*}[t]
    \centering
    \includegraphics[width=\linewidth]{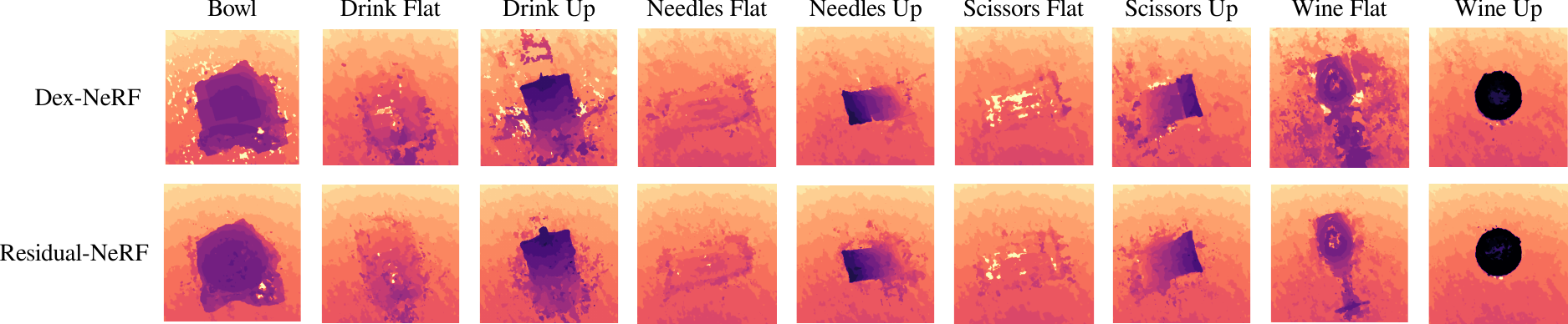}
    \caption{Depth maps for \modelname{} and Dex-NeRF evaluated on the synthetic Blender dataset. The results suggest \modelname{} improves depth maps with fewer holes and less noise.}
    \vspace{-3mm}
    \label{fig:blender_depth_maps}
\end{figure*}



\section{Experiments}
We evaluate \modelname{} against the baselines (Section~\ref{sec:baselines}) on synthetic Blender scenes (Section~\ref{sec:blender_data}) with ground truth depth. We also present a qualitative evaluation against the baselines on real-world scenes (Section~\ref{sec:real_depth}). The results suggest \modelname{} outperforms the baselines by generating higher-quality depth maps. Finally we show \modelname{} speeds up training (Section~\ref{sec:speed}), can improve grasp planning using Dex-Net (Section~\ref{sec:grasp}), and we ablate the impact of the Mixnet (Section~\ref{sec:mix_ablation}).




\begin{table*}[t]
\caption{Root Mean Square Error (RMSE)  in synthetic Blender Scenes.}
\label{tab:rmse}
\resizebox{\linewidth}{!}{%
\begin{tabular}{llllllllll}
\hline
         &  Bowl & Drink Flat & Drink Up & Needles Flat & Needles Up & Scissors Flat & Scissors Up & Wine Flat & Wine Up \\ \hline
Method   &   \multicolumn{9}{c}{RMSE $\downarrow$}        \\ \hline
NeRF~\cite{mildenhall2020nerf}     &  0.1900 & 0.6398 & 0.4361 & 0.3529 & 0.1599 & 0.5523 & 0.1622 & 0.2166 & 0.5973   \\
Dex NeRF~\cite{IchnowskiAvigal2021DexNeRF} &  0.0365 & 0.1065 & 0.0699 & 0.0293 & \textbf{0.0431} & 0.2624 & 0.0320 & 0.0425 & 0.3159    \\
\modelname{}     & \textbf{0.0213} & \textbf{0.0234} & \textbf{0.0316} & \textbf{0.0208} & 0.0437 & \textbf{0.0506} & \textbf{0.0289} & \textbf{0.0388} & \textbf{0.2462}      \\ \hline
\end{tabular}%
}
\end{table*}

\begin{table*}[]
\caption{Mean Absolute Error (MAE) in synthetic Blender Scenes.}
\label{tab:mae}
\resizebox{\linewidth}{!}{%
\begin{tabular}{llllllllll}
\hline
        & Bowl & Drink Flat & Drink Up & Needles Flat & Needles Up & Scissors Flat & Scissors Up & Wine Flat & Wine Up \\ \hline
Method   &   \multicolumn{9}{c}{MAE $\downarrow$}        \\ \hline
NeRF~\cite{mildenhall2020nerf}     &   0.1453 & 0.3424 & 0.3062 & 0.1698 & 0.1067 & 0.2505 & 0.1103 & 0.1483 & 0.2157  \\
Dex NeRF~\cite{IchnowskiAvigal2021DexNeRF} & 0.0195 & 0.0189 & 0.0255 & 0.0127 & \textbf{0.0193} & 0.0302 & 0.0160 & 0.0255 & 0.0338    \\
\modelname{}     & \textbf{0.0140} & \textbf{0.0138} & \textbf{0.0170} & \textbf{0.0115} & 0.0207 & \textbf{0.0129} & \textbf{0.0140} & \textbf{0.0142} & \textbf{0.0238}     \\ \hline
\end{tabular}%
}
\vspace{-3mm}
\end{table*}

\subsection{Hyperparameters}
\label{sec:hyperparams}
NeRFs require the setting of numerous hyperparameters potentially with great impact on performance. We have kept all vanilla NeRF parameters constant between runs, e.g., scene area, scale, and adaptive ray marching turned off. We set $m$ (Section~\ref{sec:nerf_depth}) to 3.0 for all Blender scenes evaluations for all methods, and tuned to the best of our ability for each real environment and method. We tune $m$ by starting from 0.0, to then increase $m$ until no more holes in the depth maps disappear. From this point, increasing $m$ will lead to more noise and no additional gains. 

\subsection{Baselines}
\label{sec:baselines}
As baselines, we evaluate \modelname{} against NeRF~\cite{mildenhall2020nerf} and Dex-NeRF~\cite{IchnowskiAvigal2021DexNeRF}. While other multi-view stereo (MVS) methods for transparent objects exist, to the best of our knowledge, they do not accept arbitrary poses.

\subsection{Synthetic Blender Data}
\label{sec:blender_data}
\modelname{} cannot be evaluated on existing datasets such as ClearPose~\cite{chen2022clearpose}, which captures 63 transparent objects, due to the lack of background images for training the background NeRF. Therefore, we use Blender~\cite{blender} to render nine photo-realistic objects as shown in Figure~\ref{fig:scenes}. We use the poses in a hemisphere from the original NeRF datasets~\cite{mildenhall2020nerf} to render train, test, and validation images. Background NeRF and residual NeRF receive images taken from the same 100 train poses. 



\subsection{Real World Data}
\label{sec:real_data}
We take images of transparent objects on a wooden table to evaluate \modelname{} in the real world. We take the images using a Nikon D3200 DSLR camera, the poses are taken in a random hemisphere. Lighting conditions, focal length, aperture, ISO, and white balance were constant while collecting datasets. We use COLMAP~\cite{schoenberger2016sfm} to retrieve the camera poses.

The background NeRF receives 100 images to create a strong scene prior. The residual NeRF, Mixnet, and the baselines are trained with 50 images, excluding those omitted by COLMAP. This experiment design simulates scenarios where a predominantly static scene can be extensively observed from many poses, but the scene under evaluation is limited to a smaller number of viewpoints.

The three evaluation scenes A-C in increasing difficulty are visible in Figure~\ref{fig:scenes}. Scene A has four opaque objects in the background scene, with an added transparent coffee container with coffee in the evaluation scene. Scene B has the same four opaque objects in the background scene but with a wine glass added in the evaluation scene. Scene C has 6 transparent objects present in the background in the background scene, 3 more transparent objects are added in the evaluation scene: a wine glass, kitchen wrap, and a glass bottle with a blue cap.  





\subsection{Implementation Details}
\label{sec:implementation_details}
We implement \modelname{} and the baselines by building on Torch-NGP~\cite{torch-ngp}. Torch-NGP~\cite{torch-ngp} is a PyTorch version of InstantNGP~\cite{mueller2022instant}, and uses a multi-resolution hash grid to speed up training. It runs at a similar but slightly slower speed as compared to Instant-NGP while offering a more efficient development cycle with only raytracing implemented in CUDA. We train \modelname{} and all baselines on an RTX 4090 and an AMD Ryzen Threadripper PRO 5965WX with 256\,GB of RAM.

\subsection{Blender Depth Results}
\label{sec:blender_depth_results}
We evaluate the inferred depth maps by \modelname{} and the baselines by comparing them against the ground truth provided by Blender. 

\subsubsection{Quantitative Comparison}
We compare \modelname{} against NeRF and Dex-NeRF by computing the MAE (Equation~\ref{eq:MAE}) and RMSE (Equation~\ref{eq:RMSE}). 

\begin{equation}
\label{eq:MAE}
    \text{MAE} = \frac{ \sum_{(i,\mathbf{r}) \in \Omega_r} \lVert\hat{D}_i(\mathbf{r}) - D_i(\mathbf{r}) \rVert_1 }{n},
\end{equation}

\begin{equation}
\label{eq:RMSE}
    \text{RMSE} = \sqrt{\frac{ \sum_{(i,\mathbf{r}) \in \Omega_r} \lVert\hat{D}_i(\mathbf{r}) - D_i(\mathbf{r}) \rVert^2 }{n}},
\end{equation}

Here $i \in [0,...,N]$ is the frame number, $\mathbf{r}$ is the pixel location, and $\Omega_r$ is the set of all pixel locations across frames. $\hat{D}(\mathbf{r})$ is the inferred depth in meters, $D(\mathbf{r})$ is the GT depth in meters. We crop each image before evaluation to focus evaluation on the transparent object. In each crop, the entire transparent object is visible, while the background is partially cropped out.

Table~\ref{tab:rmse} shows the RMSE and Table~\ref{tab:mae} the MAE for \modelname{} compared against the baselines. It shows that \modelname{} outperforms the baselines except for the `Needles Up' scene. 


\subsubsection{Qualitative Comparison}
Figure~\ref{fig:blender_depth_maps} shows the depth maps inferred by Dex-NeRF~\cite{IchnowskiAvigal2021DexNeRF} and \modelname{}. For most scenes, it shows that \modelname{} reduces holes and noise in depth maps. Intuitively, the background NeRF makes it less likely for holes in the table to appear, as a result of the scene geometry prior. 

The bowl, drink up, and wine flat scenes show that transparent objects can make reconstruction of opaque objects more difficult, as a result of the introduced ambiguity. Holes in opaque objects may cause the end effector to collide with unseen objects.

The results could be further improved by tuning $m$ (Section~\ref{sec:nerf_depth}) for each scene, we have opted for setting $m = 3$ for all scenes. In our testing, the optimal $m$ appears to be scene-dependent as well as NeRF implementation-dependent. Ichnowski et al.~\cite{IchnowskiAvigal2021DexNeRF} found $m = 15$ to work best, evaluating different scenes and using a NeRF implementation without multi-resolution hash encoding.

\subsection{Real World Depth Results}
\label{sec:real_depth}
\begin{figure}[t]
    \centering
    \includegraphics[width=\linewidth]{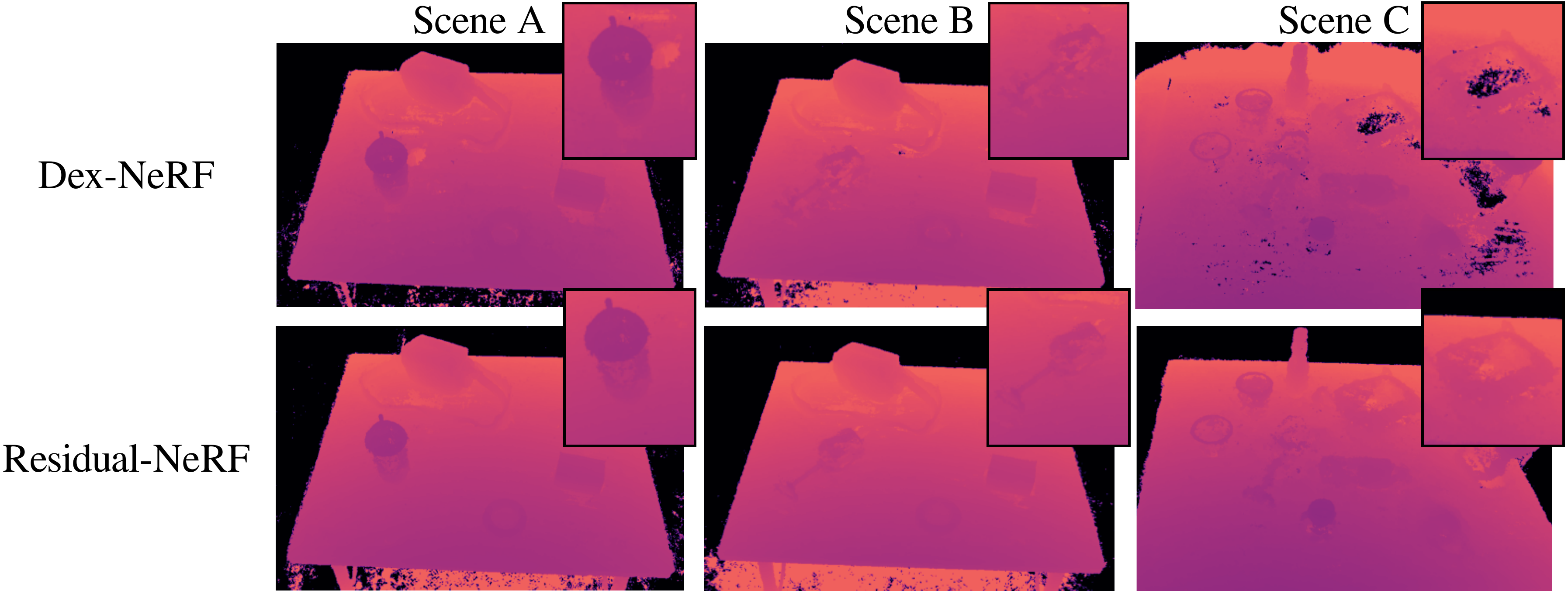}
    \caption{Depth maps inferred by Dex-NeRF~\cite{IchnowskiAvigal2021DexNeRF} and \modelname{} in the real world.\label{fig:real_data}. The result suggest \modelname{} results in fewer holes and less noise.}
    \vspace{-5mm}
\end{figure}

Figure~\ref{fig:real_data} shows the depth maps inferred by Dex-NeRF~\cite{IchnowskiAvigal2021DexNeRF} and \modelname{} in the real world. The results show \modelname{} improves the depth maps using the background NeRF as a scene prior. Similar to the results in simulation, we observe fewer holes in the table and in the objects themselves.

\begin{figure}
    \centering
    \begin{subfigure}{\linewidth}
        \includegraphics[width=\linewidth]{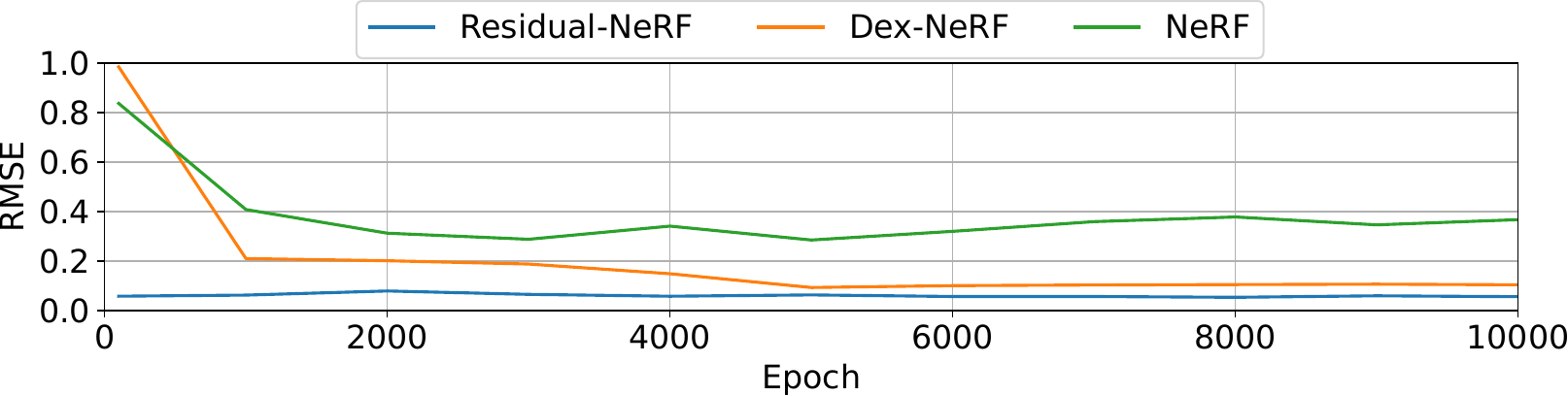}
        \vspace{-6mm}
        \caption{Root Mean Square Error (RMSE).}
    \end{subfigure}
    \begin{subfigure}{\linewidth}
        \includegraphics[width=\linewidth]{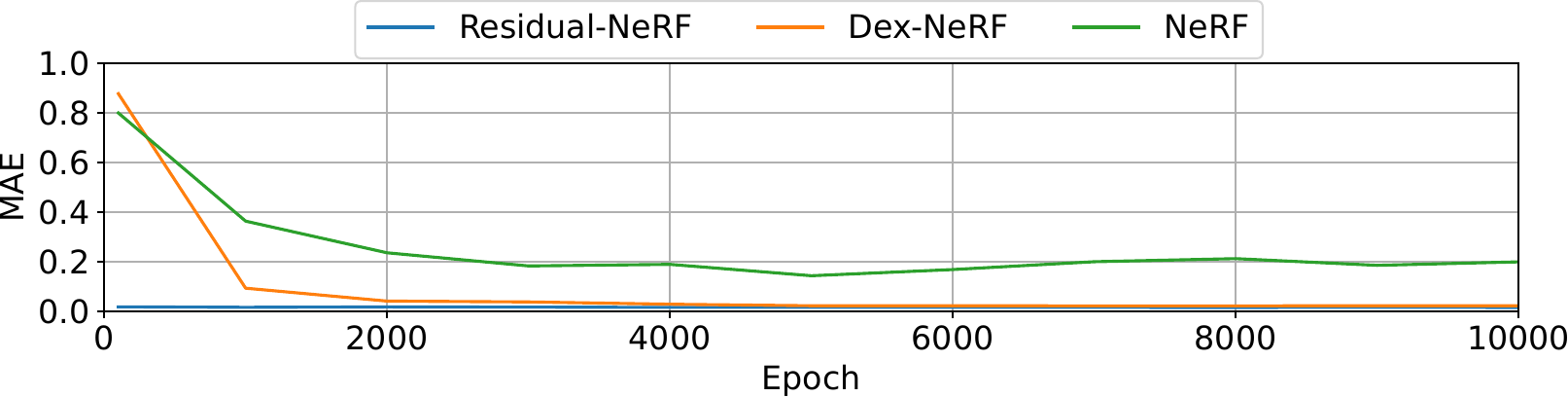}
        \vspace{-6mm}
        \caption{Mean Absolute Error (MAE).}
    \end{subfigure}
    \vspace{-4mm}
    \caption{RMSE and MAE in depth maps, logged over the number of training epochs and averaged over all synthetic scenes. The results suggest \modelname{} greatly improves training speed with respect to depth reconstruction. }
    \label{fig:speed}
\end{figure}
\subsection{Training Speed}
\label{sec:speed}
Speeding up NeRF training is critical to making NeRF a viable option in industrial and home applications. Any additional speed-ups provided by \modelname{} can further improve sped-up implementations such as Evo-NeRF~\cite{kerr2022evo}. 

To evaluate the quality of depth reconstruction over time, we log depth maps along with the elapsed training epochs. The resulting RMSE and MAE averaged over all synthetic scenes during training is visible in Figure~\ref{fig:speed}, assuming a pre-trained background NeRF for \modelname{}. 

The results suggest \modelname{} successfully leverages the background NeRF to speed up training. We implement all methods in the same Torch-NGP~\cite{torch-ngp} codebase, making implementation-related differences in training speed unlikely.

\subsection{Grasp Planning using Dex-Net}
\label{sec:grasp}
\begin{figure}[t]
    \centering
    \includegraphics[width=\linewidth]{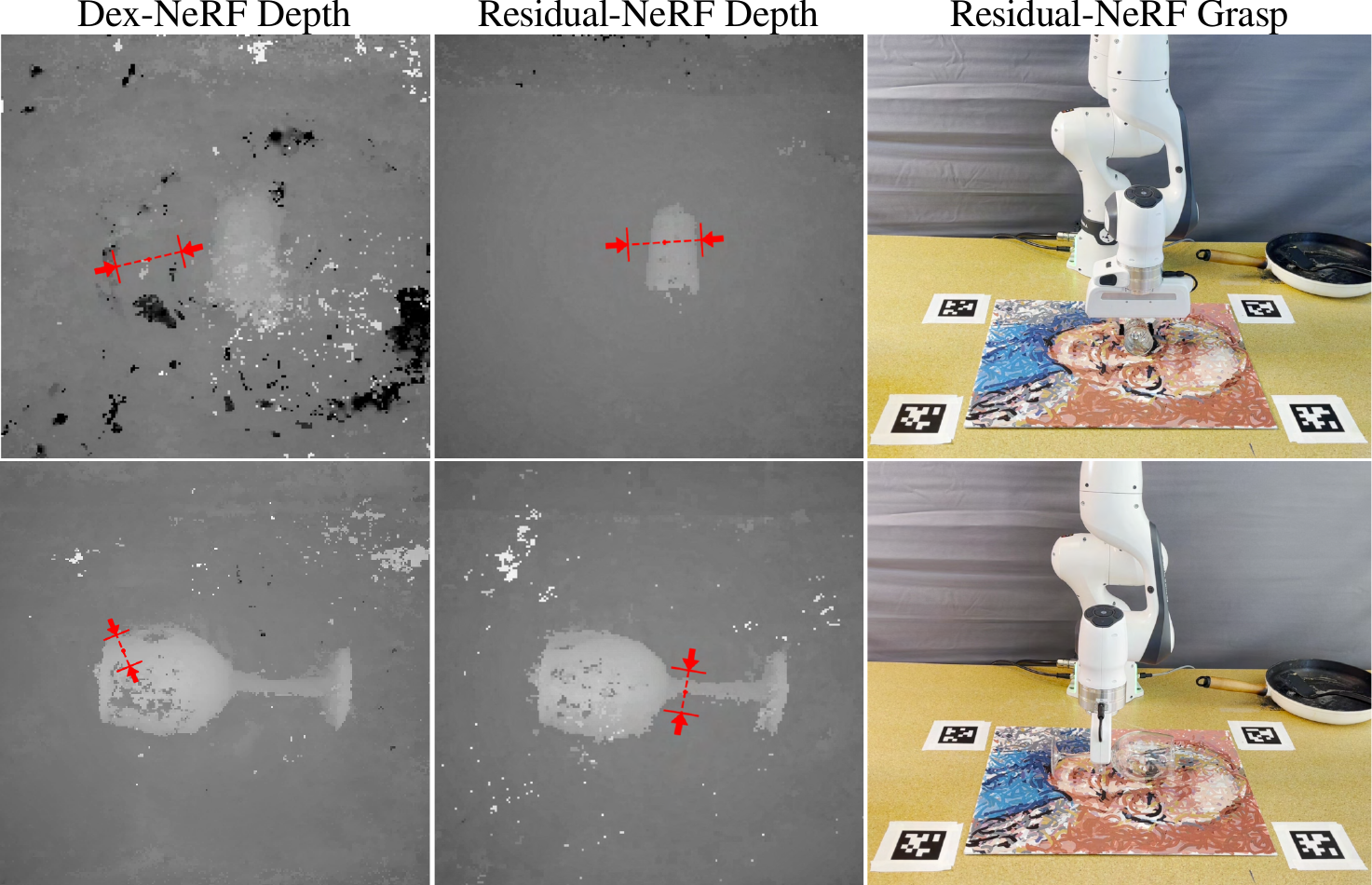}
    \caption{Dex-Net evaluated on the output of Dex-NeRF and Residual-NeRF. The results suggest the output from Residual-NeRF leads to more robust grasping.}
    \label{fig:grasp}
    \vspace{-5mm}
\end{figure}

Figure~\ref{fig:grasp} shows the output of Dex-Net on \modelname{} and Dex-NeRF. The results suggest the output from Residual-NeRF leads to more robust grasping. We use the grasp location inferred by Dex-Net to grasp transparent objects using a Franka robot manipulator and its parallel-jaw gripper.

\subsection{Residual-NeRF-S Mixnet Ablation}
\label{sec:mix_ablation}

To better understand the use of the Mixnet, we evaluate an alternative method where NeRFs are merged by adding. We refer to this method as Residual-NeRF-S. The equations for Residual-NeRF-S are:

\begin{equation}
\label{eq:residual_nerf_naive}
\resizebox{0.88\linewidth}{!}{$
\hat{C}(\mathbf{r}) = \sum_{i=1}^{N} T_i (1-\text{exp}(-(\sigma_\mathrm{bg}+ \sigma_\mathrm{res})_i \delta_i))(\mathbf{c}_{\mathrm{bg}+\mathrm{res}})_i,
$}
\end{equation}
with
\begin{align*} 
    \mathbf{c}_{\mathrm{bg}+\mathrm{res}} & = S( \mathbf{c'}_\mathrm{bg}+\mathbf{c'}_\mathrm{res}),\\
    (\mathbf{c'}_\mathrm{bg}, \sigma_{bg}) & = \Phi_\mathrm{bg} ( h(x,y,z,\theta,\phi)),  \\
    (\mathbf{c'}_\mathrm{res}, \sigma_{res}) & = \Phi_\mathrm{res} ( h(x,y,z,\theta,\phi)),  \\
    S(x) & = \frac{1}{1+e^{-x}}, 
\end{align*}

\begin{figure}[t]
    \centering
    \includegraphics[width=0.9\linewidth]{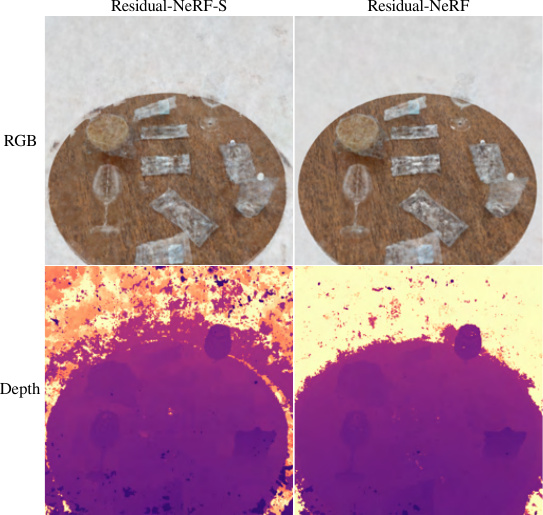}
    \caption{Naive Merging (Residual-NeRF-S) as introduced in Equation~\ref{eq:residual_nerf_naive} compared against Mixnet (\modelname{}). The results suggest Mixnet helps view reconstruction and depth mapping when merging NeRFs.}
    \label{fig:mix_ablation}
    \vspace{-5mm}
\end{figure}

Instead of training a Mixnet to merge background and residual NeRFs, the outputs of both MLPs are simply added before their activation layer. Figure~\ref{fig:mix_ablation} shows a comparison of depth and view reconstruction, both are worse compared to \modelname{}.

Especially with partially opaque objects, significant noise is introduced in the view reconstruction and depth maps. We believe this is due to the unpredictable behavior of the background NeRF in areas that were previously unoccupied. Without Mixnet, the residual NeRF not only has to understand the scene but also deduce the background NeRF's color and density predictions in previously empty regions.

\section{Conclusion and Discussion}
In this work, we introduce \modelname{}, a method designed to leverage a strong prior on the background scene to improve depth perception and speed up training. We study the problem of perceiving depth in mostly static scenes with transparent objects, as is typical of a robot's working area, e.g., shelves, desks, and tables.

\modelname{} begins by learning a \emph{background NeRF} of the entire scene without transparent objects. Following this, a \emph{residual NeRF} and \emph{Mixnet} are trained to complement the \emph{background NeRF}. The results suggest \modelname{} learns faster and achieves better depth mapping, compared to the baselines.

This work could be improved by comparing against more MVS methods non-specific to transparent objects. Future work may also include combining \modelname{} with recent advances in depth map completion. Future research could explore the performance across different transparent objects and environment conditions.

\section{Acknowledgement}
We thank the Center for Machine Learning and Health (CMLH) for their generous Fellowship in Digital Health. We also thank the Pittsburgh Supercomputing Center for their resources, and Uksang Yoo, Peter Schaldenbrand and Jean Oh for help with the robot experiments.

\bibliographystyle{IEEEtran}
\bibliography{egbib}  

\end{document}